# Neural Operator Surrogates for Two-Dimensional Neutron Flux Estimation

Japan K. Patel,[1,*] Barry D. Ganapol,[2] Anthony Magliari,[3] Matthew C. Schmidt,[4] and Todd A. Wareing[5]
[1,*] *Gateway Scripts LLC, St. Louis, MO, jpatel@gatewayscripts.com*
[2] *University of Arizona, Department of Aerospace and Mechanical Engineering, Tucson, AZ, ganapol@cowboy.arizona.edu*
[3] *Siemens Healthineers, Medical Affairs, Palo Alto, CA, anthony.magliari@siemens-healthineers.com*
[4] *Washington University in St. Louis, St. Louis, Department of Radiation Oncology, St. Louis, MO, matthew.schmidt@wustl.edu*
[5] *Siemens Healthineers, Radiation Physics, Palo Alto, CA, todd.wareing@siemens-healthineers.com*

## INTRODUCTION

Repeated transport solves form the basis of many tasks in radiation shielding and reactor analysis, including design exploration, uncertainty quantification, and optimization. The standard workflow specifies a governing equation, domain, material cross sections, sources, and boundary conditions; discretizes in angle and space; and solves iteratively to obtain the scalar flux. This procedure is accurate and interpretable, but obtaining a converged solution can become expensive, and that cost is incurred anew for every configuration when each of the many setups must be explored. In scattering-dominated regimes the difficulty is compounded, as the spectral radius of source iteration approaches unity and the iteration count grows accordingly.

Neural operators offer a complementary route. While conventional neural networks learn mappings between finite-dimensional vectors, neural operators infer mappings between function spaces. Once trained, they produce a solution for a new input with a single forward pass [1]. Fourier neural operators (FNOs) represent the operator kernel in the frequency domain using spectral-convolution layers [2]. U-shaped neural operators (UNOs) add a multiscale encoder–decoder structure with skip connections to combine global and local features [3]. In neutron transport, operator-learning surrogates have so far been demonstrated in idealized settings. For one-dimensional slab geometry, DeepONets have been trained to map a source distribution to the scalar flux in a uniform medium [4]. FNO and DeepONet surrogates have also been compared on fixed-source slabs, trained separately for each scattering-ratio regime. The FNO was more accurate, and both required well under a percent of a discrete ordinates solve [5]. In two dimensions, a DeepONet has been used to map the source distribution to the neutron flux on a fixed geometry, as a real-time surrogate for a reactor digital twin [6]. In each of these studies, the medium or geometry is held fixed while the source varies.

Machine learning has been applied to transport tasks such as iterative-solver recommendation [7] and the automated discovery of sequence acceleration schemes [8]. More recently, we used FNOs to construct AI-enhanced low-order and single-sweep solutions for one-dimensional slab-geometry transport with isotropic [9] and highly anisotropic forward-peaked scattering [10]. We also applied low-order surrogate ideas to dosimetry problems with spatial heterogeneities [11]. These efforts extend a broader class of high-order/low-order methods, in which an inexpensive low-order operator informs or accelerates the high-order transport solve [12, 13].

This work extends our one-dimensional single-sweep neural-operator studies [9, 10] to two dimensions. We consider one-group transport with isotropic scattering. As in the one-dimensional work, we use Fourier neural operators (FNOs) to approximate high-fidelity scalar flux. Additionally, we investigate U-shaped neural operators (UNOs) in this study. We consider three surrogates. The first two map the material and source fields directly to the flux, one using an FNO and one using a UNO. The third is an FNO that additionally takes the scalar flux after one source iteration, the single-sweep approximation, as an input. Each case is solved to high fidelity with a verified discrete-ordinates solver, and an average relative $L_2$ error norm is used to characterize the quality of the inferred maps. We train every surrogate over three random seeds so that differences between them can be assessed against run-to-run variability. Two questions guide the study: whether the single-sweep input improves accuracy over the direct maps, and whether training on the logarithm of the flux improves accuracy in the strongly attenuated regions relevant to shielding.

## REFERENCE TRANSPORT PROBLEM

We consider a steady state, monoenergetic transport problem on a rectangular domain $D = [0,10] \times [0,10]$. For direction $\mathbf{\Omega}$, angular flux $\psi(\mathbf{x}, \mathbf{\Omega})$, scalar flux $\phi(\mathbf{x})$, total and scattering cross sections $\Sigma_t(\mathbf{x})$ and $\Sigma_s(\mathbf{x})$, and isotropic source $q(\mathbf{x})$,

$$\mathbf{\Omega} \cdot \nabla\psi + \Sigma_t\, \psi = \frac{\Sigma_s}{4\pi}\, \phi + \frac{q}{4\pi}, \tag{1}$$

$$\phi(\mathbf{x}) = \int_{4\pi} \psi\,(\mathbf{x}, \mathbf{\Omega})\, d\mathbf{\Omega}, \tag{2}$$

with vacuum boundary conditions on all sides. Using standard notation, Eq. (1) can be written in operator form as

$$L\psi = S\psi + q, \tag{3}$$

where $L$ is the leakage plus collision operator and $S$ is the scattering operator. The angular domain is discretized with a level-symmetric $S_{12}$ quadrature and spatial variables are resolved using the diamond difference scheme [14]. The converged scalar flux $\phi$ is the high-fidelity reference that the surrogates help approximate. Each surrogate is specific to this problem class: steady-state transport with specific distribution of cross-sections and source fields defined below.

## PROBLEM SETS

We construct two sets of problems that differ in their material layout. In the first set the medium is spatially uniform. Each case has constant total cross section $\Sigma_t \in [0.5, 2.0]$ and scattering ratio $c = \Sigma_s/\Sigma_t \in [0.3, 0.95]$. The second set is heterogeneous. Each case has a uniform background with cross sections drawn from the same ranges, together with two embedded, non-overlapping regions. The first region is an absorbing square block with a low scattering ratio, a few mean free paths across. The second is a disk with independently sampled cross sections. The problem is non-multiplying and fixed source. The block represents absorbing material. In both sets, the source for each case consists of one or two isotropic Gaussian bumps. For each bump we sample the center uniformly between 0.2 and 0.8 of the domain length in each direction, the width $\sigma$ in $[0.8, 2]$, and the amplitude in $[0.5, 1.5]$. The bumps are summed to give a smooth, non-negative source.

For each sample the neural-operator input is the grid function

$$u_h = \left[\Sigma_t,\ \Sigma_s,\ q,\ g_{\mathrm{mat}},\ g_q,\ x,\ y\right], \tag{4}$$

where $g_{\mathrm{mat}} = \sqrt{|\nabla \ln \Sigma_t|^2 + |\nabla c|^2}$ is a combined material gradient, and $g_q$ is the gradient of the normalized source. The material gradient is zero in the first set, where the medium is uniform, and is nonzero along the block and disk interfaces in the second set, where it is informative. The output of each surrogate is the reference scalar flux $\phi$. The uniform set contains 1024 cases, and the heterogeneous set contains 2048. Each set is split into 80% for training and 20% for testing.

## NEURAL OPERATOR SURROGATES

The direct surrogate approximates the reference operator as $\hat{\phi}_h = \mathcal{G}_\theta(u_h)$, where $\mathcal{G}_\theta$ is an FNO [2] or a UNO [3]. The single-sweep-conditioned surrogate augments the input with a single discrete-ordinates transport sweep,

$$\hat{\phi}_h = \mathcal{G}_\theta\left(u_h,\ \phi_h^{\mathrm{SS}}\right). \tag{5}$$

Here $\phi_h^{\mathrm{SS}}$ is the scalar flux after one sweep (the uncollided flux), with the scattering source set to zero. It is computed on the same grid and costs far less than the converged solve. The single-sweep approximation shows where the source streams and where the block casts a shadow. Supplying it as an input gives the network this structure directly.

The two architectures are implemented in PyTorch [15] using the NeuralOperator library [16], with identical training settings. The FNO represents global operator kernels with four spectral-convolution layers, each retaining 16 Fourier modes per spatial dimension, with 48 hidden channels. The UNO uses six layers in a U-shaped arrangement with skip connections and a base width of 32 channels. The field is coarsened to one-quarter resolution through the first layers and refined back to full resolution through the later ones, with layer channel widths 32, 64, 64, 64, 32, 32 and Fourier modes following the same U-shape, from 16 per dimension at the outer scales down to 8 at the coarsest. FNO has about 1.36 million trainable parameters and the UNO has about 1.52 million. All models map the seven input fields (eight for the single-sweep model) to the scalar flux. Training uses the AdamW optimizer [17] at a learning rate of $10^{-3}$ with weight decay $10^{-4}$, for up to 300 epochs, with the learning rate halved on validation plateaus and early stopping. The FNO uses a batch size of 16 and the UNO a batch size of 8. Surrogates are trained with a relative $L_2$ loss evaluated on the physical flux [2],

$$\epsilon = \frac{1}{N_b}\sum_i \frac{\|\hat{\phi}_i - \phi_i\|_2}{\|\phi_i\|_2}. \tag{6}$$

where $N_b$ is the number of samples in a batch, $\hat{\phi}_i$ and $\phi_i$ are the predicted and reference scalar flux for sample $i$, and $\|\cdot\|_2$ is the $L_2$ norm over the spatial grid. Because the flux varies over several orders of magnitude, we also train a variant in which the network predicts $\log_{10}\phi$; predictions are mapped back to the physical flux for the loss and for the primary metric, so the two variants remain directly comparable while the logarithmic variant additionally resolves the low-magnitude flux.

## RESULTS

We report accuracy as the relative $L_2$ error of the scalar flux on the test set. The scalar flux varies over several orders of magnitude across the domain. It is largest in the source region and falls to small values in the block's shadow and near the vacuum boundaries. The relative $L_2$ norm is therefore governed by the high flux near the source and is insensitive to errors where the flux is small. Accurate prediction of the low flux matters for shielding, so we also report the relative $L_2$ error of $\log_{10}\phi$. This measure weights all flux magnitudes comparably. Table I reports the measured accuracy.

TABLE I. Test relative $L_2$ error

| Model | Set | flux rel-$L_2$ ($\times 10^{-3}$) |
|---|---|---|
| SS-FNO | uniform | $4.47 \pm 0.18$ |
| SS-FNO | heterog. | $6.23 \pm 0.14$ |
| FNO (direct) | uniform | $4.94 \pm 0.30$ |
| FNO (direct) | heterog. | $7.22 \pm 0.04$ |
| UNO (direct) | uniform | $6.27 \pm 0.36$ |
| UNO (direct) | heterog. | $8.50 \pm 0.17$ |

For reference, the single-sweep flux alone (with no learning) gives an average relative $L_2$ of 0.59 and 0.58 on the two sets, so every surrogate is roughly two orders of magnitude more accurate than the uncollided approximation.

Several results hold across both problem sets. Single-sweep conditioning is the most accurate formulation. It attains the lowest flux error in both sets, followed by the direct FNO and then the UNO. On the heterogeneous set the three-seed intervals are separated, so this ordering is robust to the choice of random seed. The three formulations are nonetheless comparable in magnitude. All reach flux errors of order $10^{-3}$, and single-sweep conditioning leads by a modest margin. The UNO is consistently last, as its multiscale downsampling does not overtake the FNO on these smooth-flux problems. Training on the logarithm of the flux leaves the flux error essentially unchanged. It reduces the log-flux relative $L_2$ error from roughly 0.13–0.17 to 0.047–0.054, about a factor of 2.7, for every model and set. The network becomes more accurate in the strongly attenuated regions, deep in the block's shadow and near the domain boundaries. This gain comes at little cost to accuracy near the source. Heterogeneity imposes a modest and uniform cost. Adding the block and disk raises the flux error by about a factor of 1.4 across formulations, and all models remain below $10^{-2}$. Increasing the training set improved accuracy substantially in preliminary runs. This suggests that data volume is the dominant lever in this regime.

Fig. 1 shows three representative heterogeneous cases for the single-sweep-conditioned FNO. Across roughly five orders of magnitude in flux, the prediction is visually indistinguishable from the reference. Fig. 2 isolates the effect of logarithmic training on the two error measures. The overall flux relative $L_2$ error (left) is essentially unchanged, while the log-flux relative $L_2$ error (right), a measure of accuracy in the strongly attenuated regions, decreases by a factor of roughly 2.7 for every model and set.

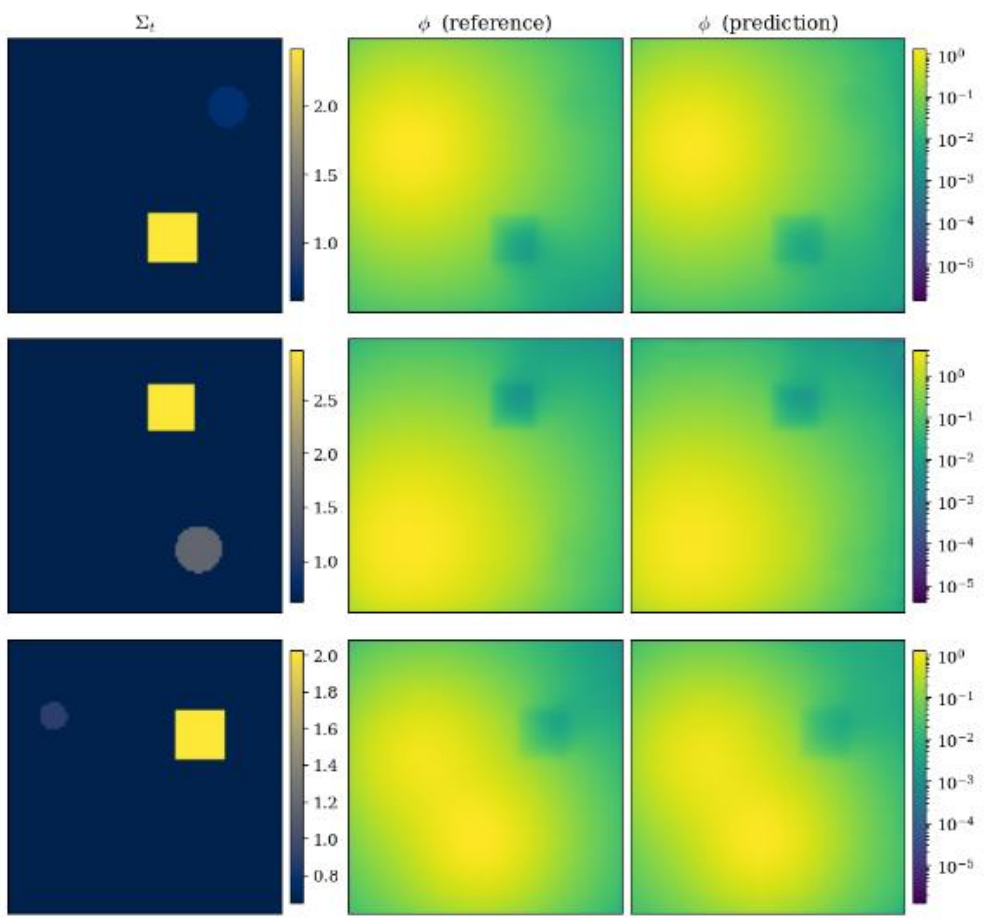

Fig. 1. Representative heterogeneous cases. Left to right: total cross section $\Sigma_t$ (uniform background, absorbing block, and disk); reference scalar flux; single-sweep-conditioned FNO prediction. Rows are three test cases spanning a wide flux range.

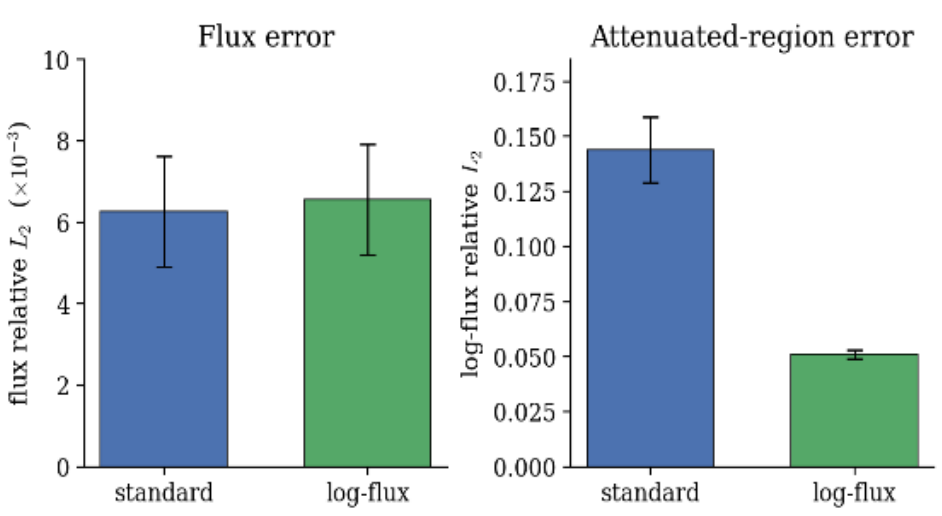

Fig. 2. Effect of logarithmic-flux training on physical-flux and log-flux relative L2 errors.

Inference and reference-solver cost run on different hardware, so we report them separately instead of as a single speedup factor. Surrogate inference was benchmarked on a single NVIDIA L4 GPU and takes about 1 ms per case. It is independent of the scattering ratio, since inference is a single forward pass with fixed computational cost. The reference $S_{12}$ solve was run on a single Intel Xeon Gold 6148 CPU (Dell workstation, 192 GB RAM) and takes 1.4 to 22 s per case. Its source-iteration count rises from about 22 to 342 as the scattering ratio increases, with a correlation of 0.84. Fig. 3 shows the two costs on their native hardware.

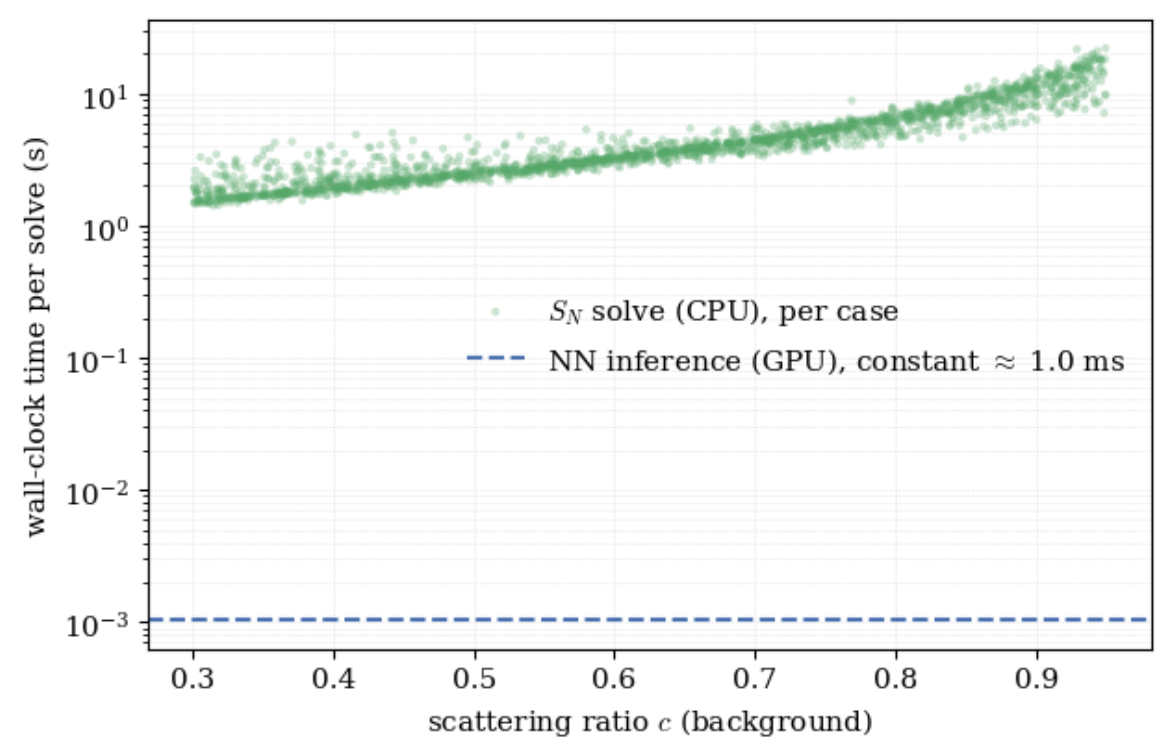

Fig. 3. Cost per solve versus background scattering ratio $c$. The two are shown on their native hardware.

## CONCLUSIONS

We formulated two-dimensional fixed-source neutron flux estimation as an operator-learning problem, in which neural operators approximated a high-fidelity discrete-ordinates reference. We compared direct FNO and UNO surrogates with a single-sweep-conditioned FNO, over two problem sets of increasing material complexity. Single-sweep conditioning was the most accurate formulation in both sets. Training on the logarithm of the flux improved accuracy in the strongly attenuated regions by roughly a factor of 2.7, at negligible cost to the overall error. This matters for shielding, where the attenuated flux far from the source governs the response. Heterogeneity from an embedded absorber imposed only a modest and uniform cost. Surrogate inference runtime was roughly constant, while the reference solves grew more expensive as the scattering ratio increased.

This study was the first step from our one-dimensional single-sweep surrogates toward multidimensional setups. The learned models gave rapid inference once data generation and training were complete. They remain tied to the training distribution and do not replace the reference solver. Future work will consider forward-peaked scattering and beam sources, along with richer heterogeneous layouts such as periodic lattices, among others. In the longer term, we will extend the formulation to energy dependence and to three dimensions.